# Subject-Event Ontology Without Global Time: Foundations and Execution Semantics

Alexander Boldachev
boldachev@gmail.com
https://orcid.org/0000-0002-7259-2952

## Abstract

A formalization of a subject-event ontology is proposed for modeling complex dynamic systems without reliance on global time. Key principles: (1) event as an act of fixation - a subject discerns and fixes changes according to models (conceptual templates) available to them; (2) causal order via happens-before - the order of events is defined by explicit dependencies, not timestamps; (3) making the ontology executable via a declarative dataflow mechanism, ensuring determinism; (4) models as epistemic filters - a subject can only fix what falls under its known concepts and properties; (5) presumption of truth - the declarative content of an event is available for computation from the moment of fixation, without external verification.

The formalization includes nine axioms (A1-A9), ensuring the correctness of executable ontologies: monotonicity of history (I1), acyclicity of causality (I2), traceability (I3). Special attention is given to the model-based approach (A9): event validation via schemas, actor authorization, automatic construction of causal chains (W3) without global time.

Practical applicability is demonstrated on the boldsea system - a workflow engine for executable ontologies, where the theoretical constructs are implemented in BSL (Boldsea Semantic Language). The formalization is applicable to distributed systems, microservice architectures, DLT platforms, and multiperspectivity scenarios (conflicting facts from different subjects).

# §1. Introduction

## §1.1. The Problem of Time and Subjectivity in Ontologies

Traditional formal ontologies (DOLCE [Masolo et al., 2003], BFO [Smith, 2016]) rely on global time as a primitive for ordering events and states. This creates fundamental difficulties:

1. **Physical infeasibility**. In distributed systems, global time synchronization is fundamentally impossible (Lamport, 1978; relativistic causality).
2. **Epistemological problem**. Actors (subjects) fix events *here and now*, without access to a global timeline. Measured time is a construct, derived from the order of discernments, not a primary given.
3. **Object-centric perspective**. Classical ontologies model the world through objects and their properties changing over time. This requires a versioning mechanism, which complicates semantics and hinders traceability of changes.

*Subjectivity as the key.* An alternative approach is the *subject-event ontology*, where:

- Event is not "something that happened in the world", but an *act of fixation of a discernment by a subject* (actor, sensor, software agent) - the ontology describes not the world "in itself", but the aggregate of discerned and fixed facts.
- A subject discerns only what falls under its models (conceptual templates). The "fence color" model allows fixing the event "fence is red"; without the model, this discernment is impossible.
- Event dependencies are fixed by the subject, not by an external observer via attachment to global time. Event $e_2$ depends on $e_1$ if the subject *explicitly indicated* this dependency.

This approach aligns with Husserl's phenomenology (internal time-consciousness), Austin/Searle's speech-act theory (an utterance as an action), and enactivism (Varela: cognition is enacted). The ontology becomes epistemically grounded: it does not describe "the world in itself", but structures the discernments of subjects.

## §1.2. The Model-based Approach: Discernment via Concepts

**Central idea**. A subject cannot fix an event of an arbitrary type. Fixation requires a *Model* - a structure that defines:

- *Semantics*: what can be discerned (concept "Fence", property "color")
- *Schema*: what values are permissible (EnumType{red, blue, green})
- *Rights*: who can fix (Permission: painter)
- *Conditionality*: what preceding events are necessary (Condition: Exists(inspection))

A Model is an *epistemic tool*, not an ontological description. Without the *M_Color* model, the subject lacks the conceptual tool to discern color, and the event "fence is red" cannot be formed. This is a fundamental difference from classical ontologies, where concepts (TBox) describe an independent world, and instances (ABox) describe its implementation.

Consequences:

1. *Ontology = Models + Events.* The History ($H$) is not a "dump of facts", but a managed space where every event is validated by a model (Axiom A9).
2. *Evolution of concepts.* Adding a new model expands the discerning capabilities of subjects. Example: introducing the "inspection report" model makes inspection events possible, which were not fixed previously.
3. *Multiperspectivity* Different subjects can use different models. Subject A discerns "color", Subject B discerns "texture". Their events coexist without global consensus (Axiom A4).

## §1.3. Causality Without Time: happens-before

Traditionally, defining causality requires referencing temporal order: "$e_1$ is the cause of $e_2$, if $e_1$ happened before $e_2$". Classical approaches use global time to determine order, which leads to the *post hoc ergo propter hoc* problem (correlation ≠ causation).

**Our approach**. Causality is a structural relation, defined by explicit references (*refs*):

> *hb := tc(refs)* (transitive closure of references)

Origin of references:

1. *Explicit* (from model Condition). Model *M_Painting* requires the presence of an inspection → the painting event references the found inspection event.
2. *Automatic* (*auto_chain* rule). For events of the same type from the same subject for the same individual, the engine (henceforth, an executable ontology engine like boldsea) automatically adds a reference to the predecessor (§6.1, requirement W3).
3. *Base.* References to the events that created the individuals (context).

Advantages:

- **No synchronization**. Each subject fixes events independently, referencing only known preceding events.
- **Traceability**. For any event *e*, its causal cone $e'|hb(e', e)$ is recoverable by traversing the graph (invariant I3).
- **Determinism**. If requirement W3 (Actor-serial per key) is met, the "last" event from a subject is uniquely determined by the maximum over *hb*, without clocks (§5.1, *ExistsMax*).

## §1.4. Presumption of Truth and Executable Ontologies

Traditional epistemology requires verification of statements before their acceptance. In distributed systems, this is either impossible (no arbiter) or expensive (consensus protocols).

**Presumption of Truth** (A5). The declarative content of an event is *available for computation from the moment of fixation*, without external verification. The system does not ask, "is it true that the fence is red?" - it accepts this as a fact fixed by the subject.

**Refutation via a new event** (A6). If it is later discovered that "the fence is not red, but green", this does not delete the old event but adds a new one: "fence is green". History matters; it is monotonic (I1), and the trace of origin is preserved.

**Executability**. Based on the history *H* and models (*guards*), the system automatically generates reactions (dataflow semantics, §6). The ontology becomes executable - it not only describes but also acts, generating new events according to rules.

## §1.5. Contribution of this Work

This work proposes:

1. *A minimalistic formalization* of a subject-event ontology without global time, where causality is defined by explicit dependencies ($hb \coloneqq tc(refs)$), and order does not require synchronization.
2. *A model-based approach as the ontology's core.* Models (A9) define the epistemic tools of subjects and validate events, but *do not create causality* - causality arises from factual references.
3. *Requirement W3 (Actor-serial per key)* as a balance between ontological purity and practical implementability: the engine ensures the determinism of the "last" value through automatic chains, without explicit slots in the models.
4. *Execution semantics* (*snapshot→batch→fixpoint*) with formal guarantees: monotonicity of history (I1), acyclicity of causality (I2), traceability (I3).
5. *Practical validation* through the boldsea system, where the axioms are implemented as architectural principles of the engine (BSL interpreter, dataflow execution, multiperspectivity).

**Applicability**. Microservices, event sourcing, DLT platforms, IoT, social networks, collaborative editing - anywhere multiple subjects fix events asynchronously without a central coordinator.

Structure of the article:

- §2: Related Work (comparison with DOLCE, BFO, Event Calculus, Actor Model, RDF)
- §3: Ontological Assumptions (D1-D5: event, subject, model, history)

- §4: Formal Model (signature, *hb*, axioms A1-A9)
- §5: Semantics of *guards* (literals, models as gate-keepers, admit policies)
- §6: Execution Semantics (auto_chain, step F(H), examples)
- §7: Invariants (I1-I3)
- §8: Discussion (scalability, LLMs, comparisons)
- §9: Conclusion

## §2. Related Work (updated version)

### §2.1. Fundamental Ontologies and the Problem of Time

DOLCE (Gangemi et al., 2002) uses a cognitive approach, distinguishing *endurants* (objects that persist through time) and *perdurants* (events that happen in time). BFO (Arp et al., 2015) adheres to a realist position, separating *continuants* (time-independent entities) and *occurrents* (temporal processes). UFO (Guizzardi, 2005) unifies structural and dynamic aspects by separating layers (core, events, agents).

**Common problem**. All three ontologies share:

1. *Dependence on time as a primitive* - events are ordered via temporal relations (before, during, after).
2. *Object-centric perspective* - events are often seen as secondary to objects ("the object changed at time t").
3. *Absence of subjectivity* - events "happen in the world", they are not "fixed by subjects". Who discerned the event and what conceptual tools were used is not modeled.
4. *Realism vs. constructivism* - it is assumed the ontology describes "the world as it is", not the fixed discernments of actors.

**Temporal Ontologies**. OWL-Time (Hobbs & Pan, 2004) formalizes moments, intervals, and Allen's relations for the Semantic Web, cementing dependence on an absolute timeline. Allen's Interval Algebra (Allen, 1983) offers 13 relations between intervals (before, meets, overlaps, etc.) but requires global time for their definition.

### §2.2. Process-Oriented Formalisms

Petri Nets (Petri, 1962) model parallelism through places, transitions, and tokens. Classical semantics assumes synchronous firing of transitions when tokens are present in input places. Timed Petri Nets add delays to edges but retain a global concept of time.

Event Calculus (Kowalski & Sergot, 1986) uses logic programming to reason about changes via *fluents* (changing properties) and events that *initiate/terminate* them. The formalism is tied to discrete time points and requires a linear order of events.

Key limitations:

- **Petri Nets**. Tokens are anonymous (no payload), no subjectivity, synchronous semantics make modeling distributed systems difficult.
- **Event Calculus**. Discrete time, complexity of verification (frame problem, ramification problem).
- **General**. Neither approach supports multiperspectivity - it's impossible to model conflicting events from different actors.

## §2.3. Timeless Events and Causality

Gustavo Romero (Romero, 2016) proposed a metaphysical theory where events are timeless primitives, from whose composition spacetime emerges. This approach radically solves the problem of time dependence at the level of fundamental physics but remains abstract and offers no executable semantics for information systems.

Timeless Events (Moss, 2021) develops the idea of events without global time for databases, using *bitemporal timestamps* (transaction time + valid time) for versioning. However, it focuses on versioning and history queries, not on causality and reactions.

Lamport's happens-before (Lamport, 1978) defines a partial order of events in distributed systems via:

- Local order of events within a process
- The "message sent → message received" relation
- Transitivity

This is the classic foundation for *logical clocks* (Lamport clocks, vector clocks) but offers no ontological semantics (what is an event? who fixes it? what concepts are involved?).

## §2.4. Actor Model and Subjectivity

The Actor Model (Hewitt, 1973) is a paradigm for concurrent computation where *actors* (independent entities) exchange asynchronous *messages*. Key principles:

- Actors encapsulate state
- Communication only via messages (no shared state)
- Asynchronicity and local decisions

Similarities with SEO:

- Subjectivity is central (actor ≈ subject)
- Asynchronicity without global time
- Local decision-making

Differences:

- No ontological semantics (what is an "event"? what concepts?)
- No formal model of causality (*hb*)
- No models as epistemic tools (actors don't discern events via concepts)
- Focus on communication, not on history and traceability

**Conclusion**. The Actor Model is SEO's closest relative regarding subjectivity, but SEO adds an ontological layer (models, causality, presumption of truth).

## §2.5. The Semantic Web: RDF, OWL, and the Problem of Statics

RDF (Resource Description Framework) represents knowledge as *triples* (subject-predicate-object). OWL (Web Ontology Language) extends RDF with logical constructs (classes, properties, restrictions).

Models in OWL:

- *TBox* (Terminological Box): classes and their relations
  (e.g., $Fence$ subClassOf $Structure$)
- *ABox* (Assertional Box): instances and their properties
  (e.g., $fence42$ hasColor red)

Key differences from SEO:

| *Aspect* | *RDF/OWL* | *SEO (this work)* |
|---|---|---|
| Models | TBox - description of the world (realism) | Models - epistemic tools of subjects |
| Triples | Static assertions | Events - acts of fixation with time and causality |
| Subjectivity | Absent (no "who asserts") | Central (actor in every event) |
| Causality | None (inference via reasoning) | $hb := tc(refs)$ - explicit graph |
| Changes | Via triple deletion/addition | Via new events (monotonicity I1) |
| Multiperspectivity | Named graphs (weak support) | A4 - conflicting facts are the norm |

**Critique**. RDF/OWL assume the TBox describes the "true structure of the world" and the ABox describes "facts about the world". In SEO, models are not descriptions but *tools of discernment*: a subject with model *M_Color* can fix a "red" event; without the model, they cannot (an epistemic, rather than an ontological, dependency).

## §2.6. Event Sourcing and CQRS

*Event Sourcing* (Fowler, 2005) is an architectural pattern where the system state is reconstructed by *replaying events* from an append-only log. CQRS (Command Query Responsibility Segregation) separates commands (changes) and queries (reads).

Similarities with SEO:

- Append-only history (I1)
- Events as primary (not objects)
- Traceability through history

**Differences**:

| *Aspect* | *Event Sourcing* | *SEO* |
|---|---|---|
| *Causality* | Order in the stream (stream position) | $hb \coloneqq tc(refs)$ - graph |
| *Subjectivity* | Weak (usually single-writer per aggregate) | Strong (actor in every event, A4) |
| *Models* | Aggregates (DDD) - business entities | Models - epistemic tools (A9) |
| *Reactions* | Command handlers + projections (imperative) | guards + dataflow (declarative) |
| *Presumption of Truth* | Not formalized | A5 - explicit axiom |

*Critique*. Event Sourcing is often tied to *single-writer semantics* (one aggregate = one writer), which hinders multiperspectivity. Causal order via stream position requires a central log (bottleneck).

## §2.7. Comparative Table

| Approach | Events | Time | Models | Subjectivity | Causality | Executability |
| --- | --- | --- | --- | --- | --- | --- |
| *SEO (this work)* | Acts of fixation | None (*hb*) | Epistemic filters (A9) | Central | *tc(refs)* | Dataflow |
| *DOLCE/ BFO* | Perdurants/ Occurrents | Global | TBox (description) | Peripheral | Temporal order | None |
| *Event Calculus* | Fluents | Discrete | Predicates | Absent | Initiate/ Terminate | Logic programming |
| *Timeless Events* | Bitemporal | Two-level | Schemas | Weak | Versioning | SQL queries |
| *Actor Model* | Messages | None (async) | None | Strong | Async delivery | Communication |
| *RDF/OWL* | Triples | None (static) | TBox/ABox (description) | Absent | None (inference) | SPARQL queries |
| *Event Sourcing* | Commands/ events | Stream position | Aggregates (DDD) | Medium | Order in stream | Projections |
| *Petri Nets* | Tokens | Sync/ Timed | Transitions | None | Firing order | Simulation |

Legend:

- **Subjectivity**. Central (every event has an actor) / Strong (actors exist) / Medium (implicit) / Weak (optional) / None
- **Executability**. Dataflow (automatic reactions) / Logic programming / Communication / Simulation / Queries (read-only)

## §2.8. Positioning of this Work

Thus, the proposed ontology is not just another model in a long line, but rather implements a *synthetic approach*, borrowing and reinterpreting key principles from related fields to solve fundamental problems of asynchronicity and multiperspectivity.

The proposed formalization *synthesizes*:

1. **From the Actor Model**. Subjectivity (actor in every event) and asynchronicity without global time.

2. **From Lamport**. Causal order via happens-before, but with ontological semantics (what an event is, what concepts are involved).
3. **From RDF/OWL**. Models as knowledge structures, but *not for describing the world*, rather as *epistemic tools* for subjects.
4. **From Event Sourcing**. Append-only history and traceability, but with *formal causality* (graph *hb* instead of stream position) and *multiperspectivity* (A4).
5. **From Timeless Events**. Rejection of global time, but with an emphasis on *reactions* (dataflow) instead of *versioning*.

Key Innovations:

- **W3 (Actor-serial per key)**. Automatic construction of causal chains without clocks and without explicit slots in models (§6.1, auto_chain).
- **A9 (Model-based generation)**. Models as gate-keepers of the ontology - an event is permissible only if a model exists that validates its semantics, authorization, and conditionality.
- **Presumption of Truth (A5) + Correction as Event (A6)**. The epistemological foundation for distributed systems without a central arbiter.
- **Dataflow semantics without time**. The *snapshot→batch→fixpoint* mechanism (§6) ensures deterministic reactions based on the causal graph, not temporal order.

**Applicability**. Existing approaches are either excessively dependent on objective time and strict causality (DOLCE, BFO, Event Calculus), offer no ontological semantics (Actor Model, Event Sourcing), or are static (RDF/OWL). The proposed formalization fills this niche - an *ontology for asynchronous, multiperspectivity, executable systems*.

Final Comparison with Key Ideas

From classical ontologies (DOLCE/BFO):

- We remove: global time, Object-centric perspective, realism
- We add: subjectivity, models as epistemic tools, presumption of truth

From the Actor Model:

- We keep: subjectivity, asynchronicity, locality
- We add: ontological semantics (models, causality, traceability)

From RDF/OWL:

- We keep: models as knowledge structures
- We change: TBox (description) → Models (epistemic tools); static → temporal (*hb*)

From Event Sourcing:

- We keep: append-only, events as primary
- We add: formal causality (graph *hb*), multiperspectivity (A4), dataflow

*Result*. A minimalistic ontology for *event-based temporal knowledge graphs*, where knowledge is not static (like RDF) and not linear (like Event Sourcing), but forms a *dynamic causal graph, managed by subjects via models*.

## §3. Ontological Assumptions

Before the formalization, we establish the basic assumptions about the ontology's structure.

**D1 (Event as an Act of Fixation)**. An event is not "something that happened in the world", but an *act of discernment and fixation* by a subject (actor, sensor, software agent). An event contains: who (actor), what was discerned (payload), about what (base), according to which model (model_id), and with explicit references to preceding events (*refs*). Without a subject, there is no event.

**D2 (Model as an Epistemic Tool)**. A Model is a structure defining:

- What the subject *can discern* (semantics: concepts, properties, actions)
- What values are *permissible* (schema: types, ranges, constraints)
- Who *can fix* (rights: roles, access conditions)
- What events are *necessary* (conditionality: *guards*, *Condition*)

A model is not a description of an independent world, but a *conceptual tool* of the subject. The model→event link is epistemic ("I know what color is"), not ontological ("color exists in the world").

**D3 (History as an Append-Only Graph)**. The History ($H$) is a monotonically growing sequence of events. Events are not deleted or modified (I1). The causality graph is defined by *refs*; timestamps (if any) do not participate in defining order.

**D4 (Subjectivity and Multiperspectivity)**. Different subjects can fix *conflicting events* about the same entity. The system does not enforce consensus (A4). Example: Sensor A fixed "temperature 8°C", Sensor B fixed "temperature 10°C". Both events coexist; their interpretation is at the business logic level (*guards*).

**D5 (Presumption of Truth)**. The declarative content of an event is *available for computation from the moment of fixation* (A5). The system does not check the "truthfulness" of the payload - the fact of fixation by a subject is sufficient. Refutations are expressed as new events (A6), not by deleting old ones.

Consequences:

- *No realism.* The ontology does not claim to describe "the world as it is", but structures the discernments of subjects, which are the only things that can influence the system's behavior.
- *No global time.* Order is causality (*hb*), defined by subjects via *refs*, not by an external observer with a clock.
- *No enforced truth.* Conflicts remain conflicts; their resolution is a policy, not an ontology.

These assumptions align with constructivist epistemology (Piaget, von Glasersfeld) and enactivism (Varela, Thompson): cognition is action, not reflection.

## §4. Formal Model of the Subject-Event Ontology

### §4.1. Event Signature

An event represents an atomic act of fixing a state change by an actor. Formally:

```
Event := { id: ID, actor: Actor, role?: Role, key: Key,
           payload: Payload, refs?: Set<EventID> }
```

Main functions:

- $id: Event \to ID$ - unique event identifier (e.g., a cryptographic hash of the content).
- $actor: Event \to Actor$ - the subject who fixed the event; a mandatory field ensuring the subjectivity of every fact (Multiperspective, history slices by subject).
- $role: Event \rightharpoonup Role$ - the actor's role in the context of the fixation (optional); determines access rights and permissions.
- $key: Event \to Key$ - a mandatory partitioning function, inducing a partition $H = \bigsqcup_k H_k$, where $H\_k := e \in H | key(e) = k$.
- $payload: Event \to Payload$ - the declarative content of the event (triple, document, metadata); its structure is defined by the model.
- $refs: Event \to Set < EventID >$ - factual references from the event to preceding events (the sole source of causality).

Formal definitions of auxiliary sets:

```
ModelID    := { m₁, m₂, ... }       (model identifiers)
Context    := { ctx₁, ctx₂, ... }   (identifiers of individuals/actions)
Key        := ModelID × Context     (partition keys)
Actor      := { a₁, a₂, ... }       (subjects)
```

```
    Role       := { r₁, r₂, ... }      (roles)
    ID         := { id₁, id₂, ... }    (unique identifiers)
    Payload    := model-dependent      (structured data)
```

Partition Key:

In a typical implementation, *key(e) = (model_id(e), context(e))*, where:

- *model_id(e): Event → ModelID* - identifier of the model used to create the event
- *context(e): Event → Context* - identifier of the entity (individual or action) the event pertains to

Formally, *Key* is the Cartesian product $ModelID \times Context$.

*History (H)* is an append-only sequence of events: $H = \langle e_1, e_2, \ldots, e_n \rangle$. For any key $k$, the set $H\_k \coloneqq e \in H | key(e) = k$ forms a partition of the history by event types.

*Note.* Any global total orders of reception (logs, *ts_sys*, logical/vector clocks) are *outside the core* as engineering options (see §8.2.1). The ontological core relies exclusively on causal dependencies via *refs*.

## §4.2. The happens-before (*hb*) Relation and Correctness Conditions

The *happens-before* (*hb*) relation defines a constructive partial order on events through factual references.

Definition of Transitive Closure:

Let $refs: Event \to Set < EventID >$ define the "direct dependency" relation. The transitive closure *tc(refs)* is defined as the smallest transitive relation containing *refs*:

$tc(refs) \coloneqq (e_1, e_2) | e_1 \in refs(e_2) \cup (e_1, e_3) | \exists e_2: (e_1, e_2) \in tc(refs) \land (e_2, e_3) \in tc(refs)$

Or, equivalently, by inductive definition:

- Base: If $e_1 \in refs(e_2)$, then $hb(e_1, e_2)$.
- Induction: If $hb(e_1, e_2)$ and $hb(e_2, e_3)$, then $hb(e_1, e_3)$.

Definition of *hb*:

  *hb := tc(refs)*

No "schema dependencies" are involved in the construction of *hb*; causality is expressed *only* through factual *refs*.

Correctness Conditions:

- **W1' (Well-formed refs, without clocks.** If $e_1 \in refs(e_2)$, then $e_1$ must already belong to H at the step $e_2$ is added (prohibits forward references to non-existent events). This is ensured by the snapshot semantics of execution.
- **W2 (Acyclic refs)**. The *refs* graph is acyclic (consequently, *hb* is a strict partial order: irreflexive, asymmetric, transitive).
- **W3 (Actor-serial per key - requirement for the engine)**. For a partition `per actor per key` (where $k = $ (model,base)), the set of events by actor *a* for key *k* forms a *chain* under *hb*: for any $e \neq e'$ from this set, $hb(e, e') \lor hb(e', e)$ holds. This requirement is implemented by the *auto_chain* rule (§6.1) and is necessary for the uniqueness of maxima in *ExistsMax* literals and the LWW policy.

*Consequence.* With W1'+W2, the *hb* relation is a strict partial order. With the addition of W3, the maximum in a `per actor per key` partition exists and is unique, ensuring the determinism of $ExistsMax(\phi, a, k)$.

*Remark.* The *snapshot→batch→fixpoint* mechanism (§6.2) pertains to execution and does not require global clocks. Causality is defined exclusively by the structure of the dependency graph.

## §4.3. Axioms of the Ontological Core

**A1 (Existence-for-system)**. An event exists for the system from the moment it is included in the history *H*; its declarative content becomes available for computation regardless of external verification.

**A2 (HB, constructive)**. The happens-before relation is defined exclusively through explicit references: `hb := tc(refs)`. No hidden causal links exist that are not expressed in the graph structure; models define the semantics and validation of events, but not causality.

**A3 (No hidden causes)**. All causes of an event are either explicitly listed in *refs* or are absent (root events). The system does not permit hidden influences not fixed in the graph.

**A4 (Multiperspective)**. Different actors can fix conflicting events about the same entity; the ontology permits the coexistence of alternative facts without global contradiction resolution. Conflicts are resolved at the policy level (§8.2) or remain unresolved.

**A5 (Presumption-of-truth)**. The declarative content of an event is available for computation from the moment of inclusion in *H*. The system does not check the "truthfulness" of the payload - the fact of inclusion in history is sufficient for its use in *guards* and *queries*.

**A6 (Correction-as-Event)**. Refutations, corrections, or cancellations are expressed as new events; past events are not deleted or modified. History grows monotonically (append-only).

**A7 (LWW via hb)**. By default, the Last-Write-Wins policy is recognized `per actor per key`: the observable value is determined by the maximum over *hb* in the set of events from actor *a* for key $k$ = (model,base). If the maximum is not unique, LWW is *undefined*. The uniqueness of maxima is ensured by requirement W3 (implementation: *auto_chain*, §6.1).

**A8 (Key-partition)**. The classifying function *key* induces a partition of history: $H = \bigsqcup_k H_k$. This ensures modularity of computations and independent processing of different event types.

**A9 (Model-based generation)**. An event $e$ is permissible for addition to $H$ only if a model $M \in Models$ exists such that:

(i) *model_id(e)* = *id(M)* (the event is created according to model M);

(ii) *actor(e)* is authorized according to *Permission(M)*;

(iii) *payload(e)* is valid according to *schema(M)*;

(iv) For every existential predicate $Exists(\phi)$ in *Condition(M)* that was satisfied on history $H$, *refs(e)* contains a reference to at least one witness event $e' \in H$ for which $matches(e', \phi)$ is true (see definition of *matches* in §5.1).

Formally for (iv):

$$\forall \big(Exists(\phi) \in Condition(M)\big): \big[\![Exists(\phi)]\!\big]_H = \text{true} \Rightarrow \exists e' \in H: matches(e', \phi) \wedge e' \in refs(e)$$

The model defines semantics and validation, but does *not* add causal edges automatically; references are added during *guard* execution (§6.2).

## §5. Semantics of guards and the Execution Mechanism

### §5.1. Literals and Formulas $\phi$

The language of formulas $\phi$ for describing *guards* is built from *base literals* and logical connectives.

Semantics of Formulas (matches function)

The function $matches: Event \times Formula \rightarrow Bool$ determines if an event $e$ satisfies a formula $\phi$, recursively:

Base Predicates:

$$matches(e, type = \tau) \Leftrightarrow type(e) = \tau \quad matches(e, field(p) = v) \Leftrightarrow payload(e)[p] = v$$
$$matches(e, has(p)) \Leftrightarrow p \in dom(payload(e))$$

where *dom(payload(e))* is the set of defined fields in the payload of event *e*.

Logical Connectives:

$$matches(e, \phi_1 \wedge \phi_2) \Leftrightarrow matches(e, \phi_1) \wedge matches(e, \phi_2) \quad matches(e, \phi_1 \vee \phi_2) \Leftrightarrow matches(e, \phi_1) \vee matches(e, \phi_2) \quad matches(e, eg\phi) \Leftrightarrow egmatches(e, \phi)$$

(only if $\phi$ is a base predicate)

For composite literals (*Exists*, *Count*, *ExistsMax*, *Order*), the semantics are defined directly on the history $H$ (see below).

Literals over History

Existence:

$$Exists(\phi) \Leftrightarrow \exists e \in H: matches(e, \phi)$$

True if at least one event satisfying formula $\phi$ exists in history $H$.

Counter:

$$Count(\phi) \geq k \Leftrightarrow |e \in H| matches(e, \phi)| \geq k$$

True if $H$ contains at least $k$ events satisfying $\phi$.

"Last" event per actor per key (without clocks, main variant):

$$ExistsMax(\phi, a, k) \Leftrightarrow \exists e \in H: actor(e) = a \wedge key(e) = k \wedge matches(e, \phi) \wedge \forall e' \in H: [actor(e') = a \wedge key(e') = k \wedge e'eqe \wedge matches(e', \phi)] \Rightarrow hb(e', e)$$

Intuitively, $ExistsMax(\phi, a, k)$ is true if, among the events from actor $a$ with key $k$ satisfying $\phi$, there exists an event $e$ that is causally-last (the maximal element in the chain ordered by the $hb$ relation).

If the maximum is not unique (parallel branches), the literal is undefined, and the *guard* does not fire. Uniqueness is ensured by W3 (see §4.2).

*Note (variant "per key")*. If necessary, a variant $MaxPerKey(\phi, k)$ can be defined as the maximum among *all* actors for key $k$. The precise formulation depends on the conflict resolution policy for parallel branches from different actors. Example policies:

**Policy 1** (deterministic choice by id):

$$MaxPerKey(\phi, k) \triangleq \exists a, e: ExistsMax(\phi, a, k) \wedge actor(e) = a \wedge key(e) = k \wedge matches(e, \phi) \wedge \forall a', e': (ExistsMax(\phi, a', k) \wedge actor(e') = a' \wedge key(e') = k \wedge matches(e', \phi)) \Rightarrow (a = a' \vee id(e) < id(e'))$$

**Policy 2** (via *ts_sys*, if available):

$$MaxPerKey(\phi, k) \triangleq \exists a, e : ExistsMax(\phi, a, k) \land actor(e) = a \land key(e) = k \land$$
$$matches(e, \phi) \land \forall a', e' : \big(ExistsMax(\phi, a', k) \land actor(e') = a' \land key(e') = k \land$$
$$matches(e', \phi)\big) \Rightarrow \big(a = a' \lor ts_sys(e) \geq ts_sys(e')\big)$$

The choice of policy is an engineering decision outside the ontological core (§8.2.1).

(**Compatibility**). For backward compatibility with previous works, we introduce the synonym *ExistsLast := ExistsMax*. "Last" now means "maximal by causality", not "by time".

Order (happens-before):

$$Order(\phi_1 \text{ BEFORE } \phi_2) \Leftrightarrow \exists e_1, e_2 \in H : matches(e_1, \phi_1) \land matches(e_2, \phi_2) \land hb(e_1, e_2)$$

True if events $e_1$ and $e_2$ exist, where $e_1$ causally precedes $e_2$ via *hb*.

Logical Connectives over Literals:

- Conjunction: $\phi_1 \land \phi_2$
- Disjunction: $\phi_1 \lor \phi_2$
- Negation: $\neg \phi$ - only over base literals (stratification)

**Stratification of Negation**. Negation $\neg$ is only permitted over base atomic predicates (*type*, *field*, *has*), but not over composite literals (*Exists*, *Count*, *ExistsMax*, *Order*). This guarantees the existence of a least fixed point when computing the batch (§6.2).

## §5.2. guards and emit

A *guard* (rule *Y*) is a pair $(G_Y, emit_Y)$:

- $G_Y$ - a formula $\phi$ over history *H*, defining the rule's activation condition.
- $emit_Y : History \rightarrow Event$ - an event constructor that generates a new event if $G_Y$ is true.

Example:

```
guard Y₁:
G_Y₁:=Exists(type='payment') ∧ ExistsMax(type='limit', Risk, client) ∧ (amount ≤ daily)
emit_Y₁ := Event{ type='exec', actor=System, ... }
```

Intuition: if a payment event exists, and the last limit from Risk for the client permits the amount, then an execution event is generated.

## §5.3. Models and Event Validation

A *Model* is a structure defining the *semantics* of permissible events:

```
Model := { id: ModelID, type: ModelType, guards: Set<guard>,
           schema: Schema, permissions: Set<(Role, Condition)> }
```

where:

- `type`: type of model (`Concept, Attribute, Relation, Action`)
- `guards`: conditions for event generation (`Condition → SetValue/SetDo`)
- `schema`: constraints on the payload (`data types, value ranges, nested properties`)
- `permissions`: roles and access conditions defining which actors can create events based on this model

Role of Models in Graph Construction:

1. *Epistemic Role.* The model defines *what an actor can discern*. Without model *M_Color*, an actor cannot fix the event "fence is red" - they lack the conceptual tool.
2. *Semantic Role.* The model defines the *event type* (*model_id* in *key*), which determines the partitions of *H* and the applicability of *guards*.
3. *Validation Role.* The model checks the event's correctness via A9 (§4.3): authorization, schema, and generation conditions.

A Model does NOT define causality directly:

The model→domain_event link is an epistemic dependency ("I know what color is"), not a causal one. Causality (*hb*) between domain events is defined by *factual references* (*refs*), which arise:

- *Automatically* (*auto_chain* for same-type events from the same actor)
- *Explicitly* (from the model's Condition, requiring the presence of other events)

Example:

```
Model M_Color:
- type: Attribute
- schema: { range: EnumType{red, blue, green} }
- permissions: { (painter, true) }
- guards: { Condition: Exists(M_Inspection, $base) }

Domain Events:
e_inspection: (fence#123, inspection, ok)    refs: [fence#123]
e_red:        (fence#123, color, red)        refs: [e_inspection, fence#123]
e_green:      (fence#123, color, green)      refs: [e_red, e_inspection, fence#123]
```

```
    hb-relations:
    hb(fence#123, e_inspection)     -  base dependency
    hb(e_inspection, e_red)         -  explicit (from Condition)
    hb(e_red, e_green)              -  auto_chain (same-type events by actor)
    hb(fence#123, e_green)          -  transitively via tc(refs)
```

*Note on refs formation.* In this example, refs(e_green) contains *all direct causes*:

- e_red (added by the *auto_chain* rule, §6.1),
- e_inspection (added from the model's Condition, as Exists(M_Inspection, fence#123) was satisfied),
- fence#123 (base dependency - the context creation event).

The reference to e_inspection is technically redundant (as e_inspection is reachable via e_red by *hb* transitivity), but it is included for *explicit conditionality* from the Condition. In optimized implementations, only a *minimal covering set* (transitive reduction of the *hb* graph) might be stored, but the semantics of *hb* do not change - the transitive closure *tc(refs)* is a identical for the minimal and full sets.

Model *M_Color* does *not create* hb(e_red, e_green) directly - the engine does that via the *auto_chain* rule (§6.1). The model only *requires* the presence of an inspection via its Condition, which adds an explicit reference to *refs* when the event is created.

## §5.4. Admission Policies (admit)

The admission policy determines which events can be added to history *H*. There are two levels of checks:

**admit_incoming(e)**. Checks an incoming event for compliance with A9:

> *admit_incoming(e)* ⇔ ∃*M* ∈ *Models*:
> *model_id(e)=M* ∧ *authorized(actor(e), M)* ∧ *valid(payload(e), M)* ∧ *refs_ok(e, M)*

**admit_Y(H)**. Checks if *guard Y* can be activated on snapshot *H*:

> *admit_Y(H)* ⇔ *[[G_Y]]_H* = *true* ∧ ∀*e* ∈ *emit_Y(H)*: *admit_incoming(e)*

This two-level structure ensures:

1. Validation of event structure (*admit_incoming*).
2. Validation of rule activation logic (*admit_Y*).

**Determinism**. Both *admit* functions are deterministic and depend only on the observed *H* and the event structure *e*, with no hidden states.

## §5.5. Examples of guards

**Example 1** - Exists + Count:

- *guard*: $Exists(type =' task') \wedge Count(type =' approval') \geq 3$
- Application: StartExecution, if a task is created and at least 3 approvals are received.

**Example 2** - ExistsMax (instead of ExistsLast):

- *guard*: $ExistsMax(type =' status' \wedge value =' closed', a = Owner, k =' task\#42')$
- Application: ArchiveTask, if the owner closed the task; uniqueness of the "last" status is ensured by W3 (implementation: *auto_chain*, §6.1).

**Example 3** - Order:

- *guard*: $Order\big((type =' create') \text{ BEFORE } (type =' assign')\big) \wedge Exists(type =' assign')$
- Application: NotifyAssignee, if the task was first created, then assigned.

**Example 4** - Negation (stratified):

- *guard*: $Exists(type =' document') \wedge eghas('deadline')$
- Application: SetDefaultDeadline, if the document is missing a deadline (negation only over the atomic predicate *has*).

# §6. Execution Semantics

## §6.1. (Non-normative) The *auto_chain* Rule

The *auto_chain* execution rule ensures W3 automatically, without explicit specification in the models.

Algorithm

When creating a new event *e* with $key = $ (model, base) from actor *a*:

1. Find the set of preceding events:
   $Prev \coloneqq e' \in H | model(e') = model \wedge base(e') = base \wedge actor(e') = a$
2. If *Prev* is not empty, find the maximal element *e_max* by *hb* in *Prev*:
   $e_{max} \coloneqq e' \in Prev: \forall e'' \in Prev: hb(e'', e') \vee e'' = e'$
3. Add *e_max* to *refs(e)*

For the first event by an actor for a (model, base) pair, *Prev* is empty, and no auto-reference is added.

Inductive Proof of Correctness

*Invariant W3(H).* For any actor $a$ and key $k = $ (model, base), the set of the actor's events for that key in history $H$ forms a chain under *hb*:

$$\forall a, k: \forall e_1, e_2 \in H: (\text{actor}(e_1) = a \land \text{key}(e_1) = k \land \text{actor}(e_2) = a \land \text{key}(e_2) = k \land e_1 \neq e_2)$$
$$\Rightarrow (\text{hb}(e_1, e_2) \lor \text{hb}(e_2, e_1))$$

*Base case.* For an empty history $H_0 = \emptyset$, the invariant $W3(H_0)$ holds trivially (no events).

*Inductive step.* Assume $W3(H)$ holds for the current history $H$. When a new event $e$ with key $k = $ (model, base) from actor $a$ is added, we must show that $W3(H \cup e)$ holds.

**Proof**:

1. Consider the set Prev = $e' \in H | \text{key}(e') = k \land \text{actor}(e') = a$ - all preceding events from actor $a$ with key $k$.
2. By the induction hypothesis $W3(H)$, the set *Prev* is a *chain* under *hb* (totally ordered).
3. **Case 1**. If *Prev* = $\emptyset$ (the first event from this actor for this key):
   a. After adding $e$, the set *{e}* is trivially a chain.
   b. *W3(H ∪ {e})* holds for *(a, k)*.

4. **Case 2**. If *Prev* ≠ $\emptyset$:
   a. Since *Prev* is a chain, a unique maximal element *e_max* by *hb* exists within it:
   $\forall e' \in \text{Prev}: \text{hb}(e', e_{\max}) \lor e' = e_{\max}$

   b. The *auto_chain* rule adds *e_max* to *refs(e)*, which yields *hb(e_max, e)* (by definition of *hb* via *tc(refs)*).
   c. For any $e' \in$ Prev:
   *hb(e', e_max)* (from the property of *e_max*)
   *hb(e_max, e)* (from *refs(e)*)
   $\Rightarrow \text{hb}(e', e)$ (by transitivity of *hb*)

   d. Therefore, the set *Prev ∪ {e}* is a chain:
   $\forall e_1, e_2 \in Prev \cup \{e\}, e_1 \neq e_2: hb(e_1, e_2) \lor hb(e_2, e_1)$

   e. $W3(H \cup \{e\})$ holds for *(a, k)*.
5. For other actors and keys, W3 is not violated, as the new event $e$ does not change the *hb*-relations between their events (only edges to/from $e$ are added).

**Conclusion**: The W3 invariant is preserved at every execution step, which ensures the uniqueness of maxima in *ExistsMax(φ, a, k)* without global time.

Thus, the *auto_chain* rule is an invariant that preserves the W3 property at every execution step, ensuring the determinism and correctness of the *ExistsMax* literal.

Origin of References

References in *refs(e)* can have different origins:

- **auto_chain**. Added by the rule above.
- **explicit**. Added from the model's *Condition* (§6.2, requirement A9(iv)).
- **base**. References to the creation events of the context (individuals).

The origin can be stored as metadata for debugging (e.g., tag *origin: auto_chain | explicit | base*), but it does not affect the semantics of *hb*. For *guard* computation, only the fact $hb(e_1, e_2)$ matters, not the reason for its truth.

## §6.2. Execution Step *F(H)*

The function *F* computes a batch of new events based on the current history *H*:

$$F(H) = H \cup B(H)$$
where $B(H) := \bigcup_{Y \in \mathcal{Y}_0} \text{emit\_Y}(H)$
and $\mathcal{Y}_0 := \{ Y \in \mathcal{Y} \mid [[G\_Y]]\_H = \text{true} \land \text{admit\_Y}(H) \}$

## Procedure

1. **Snapshot.** Fix the current history *H*.
2. **Batch Computation.** Find all active *guards* $\mathcal{Y}_0$ and generate the set of events *B(H)*.
3. **Formation of *refs* for each new event** $e \in B(H)$
   The set of references *refs(e)* is formed as the union of references from three sources:

   $\text{refs}(e) := \text{refs}_\text{explicit}(e) \cup \text{refs}_\text{auto\_chain}(e) \cup \text{refs}_\text{base}(e)$

   where:
   a. **refs_explicit(e)**. References from the model's Condition (A9(iv)). For each *Exists(φ)* $\in$ *Condition(M)* that was satisfied on *H*, a reference to at least one witness event e' $\in$ H for which *matches(e', φ)* is true is included.
   b. **refs_auto_chain(e)**. The reference to the predecessor by (model, base, actor), added by the *auto_chain* rule (§6.1). If e_max $\in$ H exists - the maximal element by *hb* among events from the same actor with the same key - then *e_max* $\in$ *refs_auto_chain(e)*. Otherwise (first event by this actor for this key), *refs_auto_chain(e) = ∅*.
   c. **refs_base(e)**. References to the creation events of the context (individuals or actions) to which event *e* pertains. For example, if `context(e) = fence#123`, then the reference to the creation event of individual `fence#123` is included in `refs_base(e)`.
4. **Update.** H' := H $\cup$ B(H).
5. **Iteration.** Repeat steps 1-4 until a fixed point is reached (B(H*) = ∅).

**Theorem (Existence of a Fixed Point).** If the stratification of negation (§5.1) is respected and the set of *guards* $\mathcal{Y}$ is finite, the process $F$ converges to a least fixed point $H^*$ in a finite number of steps.

*Proof sketch*:

- Monotonicity of F: $H \subseteq F(H)$ (append-only).
- Stratification of negation guarantees the absence of cyclic dependencies between *guards*.
- The finiteness of $\mathcal{Y}$ and the local finiteness of activation conditions (each *guard* checks a finite number of events) guarantee that in a finite number of steps, either quiescence is reached ($B(H) = \emptyset$) or all *guards* become "saturated".
- A detailed proof via fixed-point theory for monotonic operators on complete lattices is beyond the scope of this article.

## §6.3. Example: Reaction Chain

```
H₀ = { e₁: payment(amount=950, client='A') }
    refs(e₁) = []   (root event)

Step 1:
𝒴₀ = { Y₁: CheckLimit }   (active, because 'payment' exists)
B₁ = { e₂: limit_ok(client='A') }
refs(e₂) = [e₁]   (explicit from Condition Y₁: Exists(payment))
H₁ = H₀ ∪ B₁

Step 2:

𝒴₁ = { Y₂: ExecutePayment }   (active, because 'limit_ok' exists)
B₂ = { e₃: executed(client='A') }refs(e₃) = [e₁, e₂]
   where: e₁ added by auto_chain (predecessor 'payment' by (model, client, actor))
          e₂ added by explicit (from Condition Y₂: Exists(limit_ok))

H₂ = H₁ ∪ B₂

Step 3:
𝒴₂ = ∅   (no new active guards)
H* = H₂   (fixpoint)

Final hb graph:
e₁ → e₂ (explicit)
e₁ → e₃ (auto_chain)
e₂ → e₃ (explicit)
```

# §7. Correctness Invariants

**I1 (No-retro-erase / monotonicity of history).** Events are not deleted from *H*; reactions that have already fired are not annulled. New events can only initiate compensations, but not erase the past.

The creation of a new event by an actor does not cancel parallel events from other actors, even for the same key *k*. Parallel branches of history (incomparable via *hb*) coexist; their merging (if required) is implemented via aggregation policies outside the ontological core (§8.2).

**I2 (Acyclicity of *hb*).** The *hb* relation is acyclic, thanks to W1' (prohibition of forward references) and W2 (acyclicity of *refs*). Formally: no chain $e_1, \ldots, e_n$, exists where $hb(e_n, e_1)$.

**I3 (Traceability).** For any event e ∈ H, its causal cone {e' | hb(e', e)} is finite and can be reconstructed by traversing the *refs* graph. This ensures auditability and analysis of the origin of every fact.

*Note:* These invariants do not require global time; they rely on *hb*, with models defining semantics and validation, not causality. The ontology is correct in fully asynchronous, distributed systems without clock synchronization.

# §8. Discussion

## §8.1. Application and Validation

The main consequence of our approach is the ability to create *executable, reactive, and traceable systems* in asynchronous domains.

- *In AI*, the ontology can serve as a foundation for agent-based systems and simulations, where agent behavior is driven by the event flow rather than a rigid, centralized script, and the *hb*-graph can be used as a common data bus. The ability to trace the origin of any state (invariant I3) is critical for creating Explainable AI (XAI).
- *In databases and distributed systems*, the model perfectly fits the architecture of immutable logs and event sourcing. It provides a formal apparatus for ensuring data integrity and auditability without needing the complex locking mechanisms typical of mutable-state systems.
- *For computation pipelines*, such as Apache Beam, the ontology offers a semantic layer, allowing for the declarative description of complex data processing logic, including out-of-order event processing and compensation scenarios.

The proposed formalization serves as the theoretical foundation for the boldsea system (Boldachev, 2024), where axioms A1-A9 and invariants I1-I3 ensure the correctness of BSL model dataflow execution. The practical work on the system and the theoretical formalization was conducted in parallel, mutually enriching each other: industrial use cases revealed requirements for the axioms (e.g., the separation of *admit_incoming/admit_Y*, the *auto_chain* rule), and the formalization provided correctness guarantees for the engine.

The system has passed PoC testing on scenarios involving executable contracts, BPMN diagram transformation, and distributed voting. Conformance tests for the formalization are presented in Appendix D.5.

## §8.2. Limitations and Future Work

### §8.2.1. Global Time as an Engineering Option

The ontological core does not require global time or a total order of event reception (*ts_sys*). Order is defined exclusively via causal dependencies (*hb := tc(refs)*), making the ontology applicable in fully asynchronous distributed systems. However, in practical implementations, an engineering attribute *ts_sys* may be introduced for auxiliary tasks, which *does not participate in defining hb* and is semantically neutral to the system core.

Optional Use of ts_sys

In centralized or weakly distributed systems, an additional *ts_sys* field (a timestamp or a sequence number in the reception log) may be introduced for:

- **Tie-breaking**. Resolving situations with parallel maxima (when W3 is violated). For example, if two actors simultaneously created events for the same key without explicit references to each other.
- **Auditing**. Fixing the real-world time an event entered the system for regulatory requirements.
- **UI/Debugging**. Sorting events for display to the user in chronological order.

*Important: ts_sys* does not participate in the definition of *hb* and does not influence the semantics of *guards*. It is an engineering attribute, not part of the ontology. The engine may use *ts_sys* for conflict resolution policies, but this happens *after hb* is computed, not instead of it.

Policies for W3 Violations

If, for a pair *(a, k)*, multiple *hb*-maximal events exist (parallel, causally incomparable branches), the system can:

1. **Conservatively**. Consider *ExistsMax(φ, a, k)* undefined (the *guard* does not fire until the conflict is explicitly resolved).
2. **Deterministically.** Choose one event based on a system attribute:
    a. Lexicographically by *id* (deterministic, but arbitrary)
    b. By *ts_sys* (if available and synchronized)
    c. By vector clocks (for distributed systems)

3. **Escalate**. Create a *conflict_detected* event for explicit resolution by an actor or the system (a system actor). For example, in a banking system, parallel limits from Risk require a manager's decision.

The choice of policy is an engineering decision outside the ontological core, dependent on domain requirements (finance requires conservatism, social networks prefer liveness).

**Relation to boldsea.** The current version of boldsea uses a `Timestamp` field, but it does not define the execution order. Order is determined only via `Cause` (≈ *refs* in SEO). `Timestamp` serves auditing and UI purposes. The SEO formalization proves the possibility of a fully asynchronous implementation without `Timestamp` as part of the semantics.

§8.2.2. Scalability and Distributed Systems

**Decentralization via Partitions:**

Partitioning history by keys (A8: $H = \bigsqcup_k H_k$) provides a natural partition for scaling:

- Events for different keys can be processed independently
- *Guards* that do not use cross-key literals are computed locally
- Replication by key subsets (*sharding*)

Scalability Limitations

1. **Cross-key dependencies**. *Guards* of the form *Exists($k_1$) ∧ Exists($k_2$)* require coordination between partitions. Minimizing such dependencies is a recommendation for model design.
2. **Causal cone size**. For an event *e* with a deep causal cone *{e' | hb(e', e)}*, validation can be costly. Optimization strategies:
    a. Caching intermediate *guard* results
    b. Incremental validation (checking only new graph edges)
    c. Checkpoints (periodic "snapshots" of a validated state)
3. **Consensus under parallelism**: In P2P networks without a central coordinator, the problem of consensus on parallel maxima arises. Solutions:
    a. CRDTs (Conflict-free Replicated Data Types) as a merge policy
    b. Byzantine fault tolerance for financial applications
    c. Eventual consistency with compensations

Future Work

- Formal verification of safety and liveness properties for dataflow models via model checking
- Integration with DLT platforms for cryptographic traceability and decentralized consensus

- Query optimization by indexing the *hb*-relation (e.g., materializing the transitive closure)

§8.2.3. Comparison with Alternative Approaches

**Event Sourcing (ES):**

*Similarities*:

- Append-only event history (I1)
- State reconstruction via event replay

*Differences*:

- SEO: causal order via *hb* (explicit graph), ES: order via stream position
- SEO: dataflow reactions (*guards*), ES: command handlers + projections
- SEO: multiperspectivity (A4), ES: usually single-writer per aggregate

**Petri Nets:**

*Similarities*:

- Tokens (≈ events), transitions (≈ *guards*)
- Causal order via firing order

*Differences*:

- Petri: tokens are anonymous, SEO: events with rich payload and actor
- Petri: global state (marking), SEO: local history *H*
- Petri: verification complexity (reachability undecidable), SEO: monotonicity simplifies analysis

**Timeless Events (Moss 2021):**

*Closest similarity, but key differences*:

- Timeless: accent on bitemporal timestamps, SEO: complete rejection of time
- Timeless: focus on versioning, SEO: focus on causality and reactions
- SEO adds: models (A9), multiperspectivity (A4), dataflow (*guards*)

**Semantic Web / RDF:**

*Similarities*:

- Triples (subject-predicate-object) ≈ events in SEO
- Ontologies (OWL) ≈ models (§5.3)

*Differences*:

- RDF: static assertions, SEO: temporal events with *hb*
- RDF: SPARQL for queries, SEO: *guards* for reactions
- SEO: presumption of truth (A5), RDF: requires entailment reasoning

## §9. Conclusion

The proposed formalization of a subject-event ontology without real time presents a minimalistic approach to modeling complex systems through events and causality. Key results:

*Theoretical Contribution:*

1. **Elimination of global time from the core**. Event order is defined exclusively via causal dependencies (hb := tc(refs)), making the ontology applicable in fully asynchronous distributed systems.
2. **Minimal set of primitives**. Event, actor, references (*refs*), models - sufficient to express complex business logic without additional abstractions.
3. **Formalization of the presumption of truth**. Axiom A5 codifies the epistemological principle of "truth through fixation", distinguishing SEO from verification logics.
4. **W3 as an engine requirement**. Separating ontological axioms (A1-A9) from execution requirements (W3, *auto_chain*) ensures theoretical purity and implementation flexibility.

*Practical Applicability:*

5. **Validation via boldsea**. The boldsea system demonstrates that the formalization is not only theoretically rigorous but also practically implementable. Proof-of-concept on real-world scenarios (contracts, BPMN conversion, voting) confirms the correctness of the axioms.
6. **Dataflow execution without clocks**. The *snapshot→batch→fixpoint* mechanism (§6) operates without global synchronization, which is critical for microservice architectures and edge computing.
7. **Multiperspectivity (A4)**. Support for conflicting facts from different actors without enforced consensus is a unique property distinguishing SEO from traditional DBMSs and Event Sourcing.

*Philosophical Foundation:*

8. **Event as an act of fixation**. Connection to speech-act theory (Austin, Searle) and phenomenology (Husserl) - truth is generated by action, not by correspondence to external reality.
9. **Correction as a new event (A6)**. Monotonicity of history (I1) aligns with the intuition of time's irreversibility and the preservation of origin traces.

10. **Models as epistemic tools**. A model is not a description of the world, but a tool of discernment for an actor. This aligns SEO with constructivism and enactivism.

Limitations and Future Work

The formalization does not cover:

- Probabilistic models (events with uncertainty)
- Temporal operators (always, eventually) - require extending the φ language
- Formal verification of *guards* (model checking for dataflow)
- Query optimization strategies for large graphs

Research directions:

- Integration with category theory for model composition
- Extension to quantum events (superposition, entanglement)
- Application to social systems (reputation, trust, actor weights)
- Connection to process mining and automatic model extraction from logs

Final Thesis

Subject-Event Ontology demonstrates that *time is not a necessary primitive for modeling dynamic systems*. Causality (*hb*), subjectivity (*actor*), and the presumption of truth (A5) are sufficient for building executable ontologies with formal guarantees of correctness. This opens the way for a new class of systems - *event-based temporal knowledge graphs*, where knowledge is not static (as in RDF) and not linearly ordered (as in Event Sourcing), but forms a dynamic causal graph, managed by actors.

# Appendices

## Appendix A. The φ Mini-language

```
φ ::= atom | φ ∧ φ | φ ∨ φ | ¬atom

atom ::= type = τ
       | field(p) = v
       | has(p)
       | Exists(φ)
       | Count(φ) ≥ k
       | ExistsMax(φ, a, k)
       | Order(φ₁ BEFORE φ₂)

where:
- τ ∈ Types  (event types)
- p ∈ Paths  (paths in payload)
- v ∈ Values (field values)
- a ∈ Actors (actors)
- k ∈ Keys = ModelID × Context (partition keys)
```

Semantics of Base Predicates (via matches):

```
matches(e, type = τ)       ⇔  type(e) = τ
matches(e, field(p) = v)   ⇔  payload(e)[p] = v
matches(e, has(p))         ⇔  p ∈ dom(payload(e))
```

Semantics of Composite Literals (on history H):

```
[[Exists(φ)]]_H          ⇔  ∃e ∈ H: matches(e, φ)
[[Count(φ) ≥ k]]_H       ⇔  |{e ∈ H | matches(e, φ)}| ≥ k
[[ExistsMax(φ,a,k)]]_H   ⇔  (см. §5.1)
[[Order(φ₁ BEFORE φ₂)]]_H ⇔ (см. §5.1)
```

*Stratification of Negation*: ¬ is applicable only to base predicates (*type, field, has*), but not to literals (*Exists, Count, ExistsMax, Order*).

## Appendix B. Manual Control Patterns (Non-normative)

In some scenarios, it may be necessary to disable *auto_chain* and manage chains manually:

**Pattern 1: Breaking the chain on context change**

```
Model M_Price:
  Condition: ($.vendor = $.prev_price.vendor) ∨ ($$.price = undefined)
```

If the vendor changes, the chain is broken (the new price does not reference the old one from a different vendor).

**Pattern 2: Explicit branching**

```
Model M_Fork:
  Condition: $.branch_point ≠ undefined
  Refs: [$.branch_point]   # Explicit ref, ignores auto_chain
```

For creating alternative scenarios (e.g., A/B testing).

*Note*: These patterns are outside the scope of the ontology and are engineering techniques. The base formalization assumes *auto_chain* is always active to comply with W3.

## Appendix C. Notes on Practices

**C.1. Query Optimization**

For large graphs (>$10^6$ events), it is recommended to:

- Materialize *hb* via indices (e.g., skip lists for transitive closure)
- Cache *guard* results for recurring patterns
- Use lazy evaluation of literals (short-circuiting for conjunctions)

**C.2. Handling Large Payloads**

If the payload contains binary data (files, images):

- Store only a hash/reference in the event
- Store the full content in a separate blob store
- The event guarantees integrity (hash matches content)

**C.3. Audit and Compliance**

For regulatory requirements (GDPR, SOX):

- Use *ts_sys* as an audit timestamp (does not affect *hb*)
- Use cryptographic signatures for events (*signature* field in payload)
- Use an immutable log for forensics